\documentclass[conference]{IEEEtran}
\IEEEoverridecommandlockouts

\usepackage{graphicx}
\usepackage{amssymb}
\usepackage[cmex10]{amsmath}
\usepackage[linesnumbered,ruled,vlined]{algorithm2e}
\usepackage{multirow} 
\usepackage{multicol} 
\usepackage{arydshln}
\usepackage{graphicx}
\usepackage{epstopdf}
\usepackage{svg}
\usepackage{booktabs}
\usepackage{mdwmath}
\usepackage{mdwtab}

\def\BibTeX{{\rm B\kern-.05em{\sc i\kern-.025em b}\kern-.08em
    T\kern-.1667em\lower.7ex\hbox{E}\kern-.125emX}}
\begin{document}

\title{STTS-EAD: Improving Spatio-Temporal Learning Based Time Series Prediction via Embedded Anomaly Detection
}

\author{\IEEEauthorblockN{1\textsuperscript{st} Yuanyuan Liang}
\IEEEauthorblockA{\textit{East China Normal University} \\
leonyuany@stu.ecnu.edu.cn
}
\and
\IEEEauthorblockN{2\textsuperscript{nd} Tianhao Zhang}
\IEEEauthorblockA{\textit{East China Normal University} \\
51215903068@stu.ecnu.edu.cn
}
\and
\IEEEauthorblockN{3\textsuperscript{rd} Tingyu Xie}
\IEEEauthorblockA{\textit{Zhejiang University} \\
tingyuxie@zju.edu.cn}
}

\maketitle

\begin{abstract}
Handling anomalies is a critical preprocessing step in multivariate time series prediction. However, existing approaches that separate anomaly preprocessing from model training for multivariate time series prediction encounter significant limitations. Specifically, these methods fail to utilize auxiliary information crucial for identifying latent anomalies associated with spatiotemporal factors during the preprocessing stage. Instead, they rely solely on data distribution for anomaly detection, which can result in the incorrect processing of numerous samples that could otherwise contribute positively to model training.
To address this, we propose STTS-EAD, an end-to-end method that seamlessly integrates anomaly detection into the training process of multivariate time series forecasting and 
aims to improve \underline{S}patio-\underline{T}emporal learning based \underline{T}ime \underline{S}eries prediction via \underline{E}mbedded \underline{A}nomaly \underline{D}etection.
Our proposed STTS-EAD leverages spatio-temporal information for forecasting and anomaly detection, with the two parts alternately executed and optimized for each other. To the best of our knowledge, STTS-EAD is the first to integrate anomaly detection and forecasting tasks in the training phase for improving the accuracy of multivariate time series forecasting. 
Extensive experiments on a public stock dataset and two real-world sales datasets from a renowned coffee chain enterprise show that our proposed method can effectively process detected anomalies in the training stage to improve forecasting performance in the inference stage and significantly outperform baselines.
\end{abstract}


\section{Introduction}
%
%
%
%
Multivariate time series (MTS) forecasting focuses on accurately predicting time series data comprising multiple interrelated variables and plays a vital role in various industries, including power consumption forecasting \cite{ shao2020domain}, stock price prediction \cite{chen2021stock}, and sales forecasting \cite{weng2020supply}.
MTS comprises multiple univariate time series, each representing a metric from a specific entity, and involves temporal dependencies within each series and spatial dependencies among series \cite{li2021multivariate}. The future of a variable depends on its own history and the combined histories of others. Compared to univariate forecasting, MTS forecasting is more complex due to spatio-temporal dependencies, dynamic changes, and noise. Temporal dependency captures patterns over time, reflecting how history shapes the future, improving trend and fluctuation modeling. Spatial dependency reveals interactions and correlations among variables, enhancing accuracy. Addressing both dependencies is crucial for effective MTS forecasting \cite{deng2021st}.

Most models struggle to address both temporal and spatial dependencies effectively. To overcome this, we propose a spatio-temporal learning-based time series prediction model (STTS), which builds spatiotemporal embeddings for each series and learns spatiotemporal features, improving prediction accuracy and reliability. STTS is highly flexible, adapting to changes in feature dimensions in MTS tasks. For example, in sales forecasting, where store sales data form the MTS, the number of stores may vary due to openings or closures, altering data dimensions. Unlike traditional deep learning models requiring retraining, STTS efficiently handles such changes, enhancing practical use. While most state-of-the-art MTS models rely on deep neural networks (DNNs) to capture complex patterns, their performance often suffers from training data anomalies and the low signal-to-noise ratio of time series data. Addressing these anomalies is critical to improving the accuracy and robustness of MTS prediction models.

\begin{figure}[t]
    \centering
    \includegraphics[width=\linewidth]{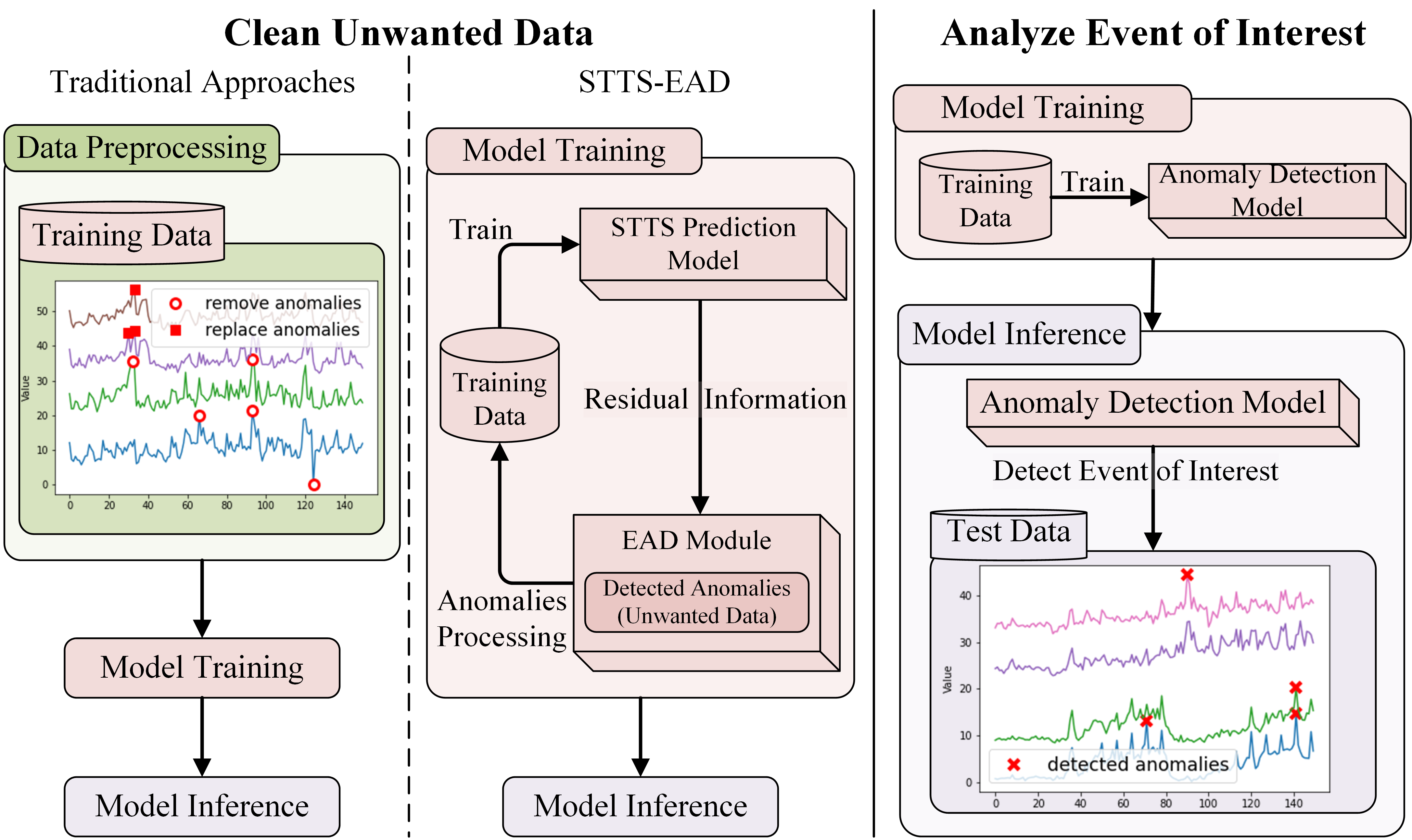}
    \caption{Taxonomy of anomaly detection methods for different purposes and positioning of STTS-EAD method.}
    \label{fig:taxonomy}
\end{figure}

Accurately handling anomalies in training data is challenging. Anomaly detection methods fall into two categories: "cleaning unwanted data" and detecting "events of interest," as illustrated in Figure \ref{fig:taxonomy}. The latter focuses on detecting events like fraud or faults by training models to learn normal sample distributions and using residuals to identify anomalies during inference. 
In contrast, data cleaning methods detect anomalies during preprocessing, removing or filling them to improve the training set. However, these methods treat anomaly detection and model training as separate stages, limiting their effectiveness. They often rely on traditional techniques \cite{li2016quality, mehrang2015outlier} that detect anomalies based on significant deviations from normal ranges, ignoring spatiotemporal factors. This omission increases false positives, leading to the unnecessary removal of useful training samples.

To address this, we propose an end-to-end approach called STTS-EAD, which integrates anomaly detection into the training process for multivariate time series forecasting, enhancing \underline{S}patio-\underline{T}emporal learning based \underline{T}ime \underline{S}eries prediction via \underline{E}mbedded \underline{A}nomaly \underline{D}etection.
More specifically, we first propose a novel \underline{S}patio-\underline{T}emporal learning based \underline{T}ime \underline{S}eries prediction model (STTS), which explicitly constructs temporal and spatial embeddings for each time series and conducts spatio-temporal feature learning to obtain more accurate and reasonable prediction results. 
Unlike conventional MTS forecasting approaches that require a fixed number of time series, our model can easily adapt to dynamic scenarios with varying numbers of time series.
Furthermore, in order to effectively address the issue of anomalies, we introduce an \underline{E}mbedded \underline{A}nomaly \underline{D}etection (EAD) module integrated into the model training process to enhance MTS prediction performance. As in Figure \ref{fig:taxonomy}, the EAD module operates during the model training stage, calculates the anomaly scores, and detects the anomalies according to the residual information from the spatio-temporal learning-based prediction model. 

The main contributions are summarized as follows:

\begin{itemize}
    \item We introduce the STTS model, which uses spatio-temporal learning for multivariate time series prediction. It constructs temporal and spatial embeddings for each time series and learns spatio-temporal features separately to improve accuracy. 
    \item We propose STTS-EAD, an end-to-end approach that enhances time series prediction with an embedded anomaly detection module. This is the first framework to integrate anomaly detection into forecasting.
    \item We validate our method on a public stock dataset and two real-world sales datasets from a multinational coffee chain. Experimental results show that STTS-EAD performs best in time series forecasting and improves prediction through anomaly detection. 
\end{itemize}

\section{Related Work}
\subsection{Time Series Forecasting}
Time Series Forecasting involves univariate and multivariate methods. Univariate methods analyze and forecast individual time series separately, offering simplicity and interpretability. 
ARIMA\cite{box2015time}, a widely used univariate method, combines autoregression and moving averages, performing well on stationary data. However, for more complex forecasting tasks with multiple variables, univariate methods fail to capture complex relationships and may not provide accurate results. These methods cannot explore the correlations between variables.
In contrast, multivariate methods consider multiple time series simultaneously, modeling both intra- and inter-metric dependencies. This allows for better prediction accuracy by capturing the dynamic relationships between series. Research on multivariate time series prediction is growing, with neural network-based methods like RNNs, CNNs, Transformers, MLPs, and GNNs emerging.

RNNs are designed for sequence data, retaining historical information for time series prediction. LSTM models improve on RNNs by using memory cells and gates to manage information flow, addressing issues like vanishing gradients. GRU models simplify LSTMs by combining memory cells with reset and update gates. Bi-RNNs \cite{schuster1997bidirectional} process sequences in both directions, capturing dependencies from past and future data .

CNNs, initially for image processing, are now adapted for time series forecasting. Dilated convolutions improve sequence pattern recognition. TCN \cite{bai2018empirical} and WaveNet \cite{oord2016wavenet} use dilated convolutions and residual connections to capture long-term dependencies. LSTNet \cite{lai2018modeling} and ConvLSTM \cite{shi2015convolutional} combine CNNs and RNNs for both short- and long-term dependencies, while TimesNet \cite{wu2022timesnet} uses Fourier transforms for multi-periodic modeling.

Transformers \cite{vaswani2017attention}, designed for NLP, are applied to time series forecasting. Models like Autoformer \cite{wu2021autoformer}, Preformer \cite{du2023preformer}, and FEDformer \cite{zhou2022fedformer} enhance learning with decomposition and improved attention mechanisms. Autoformer introduces self-correlation, Preformer uses Multi-Scale Segment Correlation (MSSC), and FEDformer employs Fourier transforms. ETSformer improves accuracy with Exponential Smoothing Attention (ESA), while Informer \cite{zhou2021informer} reduces complexity with ProbSparse attention. Pyraformer \cite{liu2021pyraformer} uses pyramid-style attention for hierarchical transmission.

Recently, \cite{zeng2023transformers} proposes using simple models like MLPs to approximate the complexity of Transformer-based models for time series prediction. MLP-based models, including NLinear, which combines time series decomposition with linear layers, and DLinear, which uses linear layers with simple data normalization, have recently achieved state-of-the-art performance \cite{zeng2023transformers}. LightTS \cite{zhang2022less} introduces continuous and interval sampling for capturing short and long-term patterns, especially in lengthy sequences. MTS-Mixer \cite{li2023mts} decomposes time and channel mixing to capture dependencies separately. TiDE \cite{das2023long} employs MLPs to encode past series and covariates, then decode for future predictions.

Lastly, GNNs enhance spatial understanding in multivariate time series data, improving predictions. MTGNN \cite{wu2020connecting} models time series as graphs, using GNNs to learn dynamic relationships for accuracy. StemGNN \cite{cao2020spectral} combines time series with graphs, capturing frequency and temporal relationships. Graph WaveNet \cite{wu2019graph} integrates GNNs and temporal convolutional networks for precise predictions. ST-GCN \cite{yan2018spatial} performs convolutions in spatial and temporal dimensions, suitable for various prediction tasks.

\subsection{Time Series Anomaly Detection}
Time Series Anomaly Detection methods can be divided into two categories according to the purpose of detection \cite{blazquez2021review}. Research is active on the purpose of the event of interest, and this family is divided into prediction-based and reconstruction-based approaches. Hundman et al. \cite{hundman2018detecting} used an LSTM-based model for spacecraft anomaly detection, introducing an adaptive non-parametric dynamic thresholding method. Munir et al. \cite{munir2018deepant} proposed DeepAnT, employing a CNN-based model to detect various anomalies in multivariate time series. CHMM \cite{zhou2016online} captures time dependencies and variable correlations incrementally. Shipmon et al. \cite{shipmon2017time} combined deep learning and statistics for anomaly detection in streaming data, utilizing rules based on predicted and actual values, tailored to each data stream.
While AE \cite{malhotra2016lstm} and VAE \cite{su2019robust} are widely employed in reconstruction-based anomaly detection. EncDec-AD \cite{malhotra2016lstm} utilizes an LSTM-based encoder-decoder architecture tailored for multivariate time series anomaly detection. It learns normal time series reconstruction patterns and detects anomalies through reconstruction errors. MSCRED \cite{zhang2019deep} combines multi-scale feature matrices, ConvLSTM networks, and convolutional decoders with residual matrices for anomaly detection, root cause analysis, and severity interpretation. OmniAnomaly \cite{su2019robust} uses techniques like random variable coupling and normalization flow to reconstruct inputs, capturing normal patterns and detecting anomalies with reconstruction probabilities for robust explanations. 
MTAD-GAT \cite{zhao2020multivariate} jointly optimizes predictive and reconstructive models, improving multivariate time series representations for anomaly detection. Alternative methods include AnoGAN \cite{schlegl2017unsupervised}, which uses DCGAN for unsupervised anomaly detection in medical images, and MAD-GAN \cite{li2019mad}, which adapts AnoGAN for time series data with an LSTM-based GAN and a novel DR-score for anomaly detection.
TAnoGAN \cite{bashar2020tanogan} maps time series data into latent space and reconstructs it through adversarial training.
USAD \cite{audibert2020usad} trains an encoder-decoder architecture with adversarial training to amplify reconstruction errors in input data containing anomalies, offering higher stability than GAN-based methods.
DAEMON \cite{chen2021daemon} employs adversarial training with two discriminators on an autoencoder to detect anomalies through reconstruction errors, ensuring robustness.

Current methods for preprocessing anomalies in training data are limited, often relying on statistical techniques like z-score \cite{li2015outlier} or box plots \cite{li2016quality} to assess deviations from the overall distribution. However, these methods may not be effective for time series data due to its seasonal and cyclical nature. For time series anomaly detection, most methods focus on individual time series. Simple approaches use constant or piecewise constant models with sliding windows and median-based references \cite{basu2007automatic, mehrang2015outlier}. Smoothing techniques like B-spline \cite{chen2010automated}, EWMA (Exponentially Weighted Moving Average) variants \cite{carter2012probabilistic}, and methods such as SCREEN \cite{song2015screen} and those by Zhang et al. \cite{zhang2016sequential} employ slope constraints for streaming data, allowing for rapid anomaly detection and correction.

\begin{figure*}
    \centering
    \includegraphics[width=\linewidth]{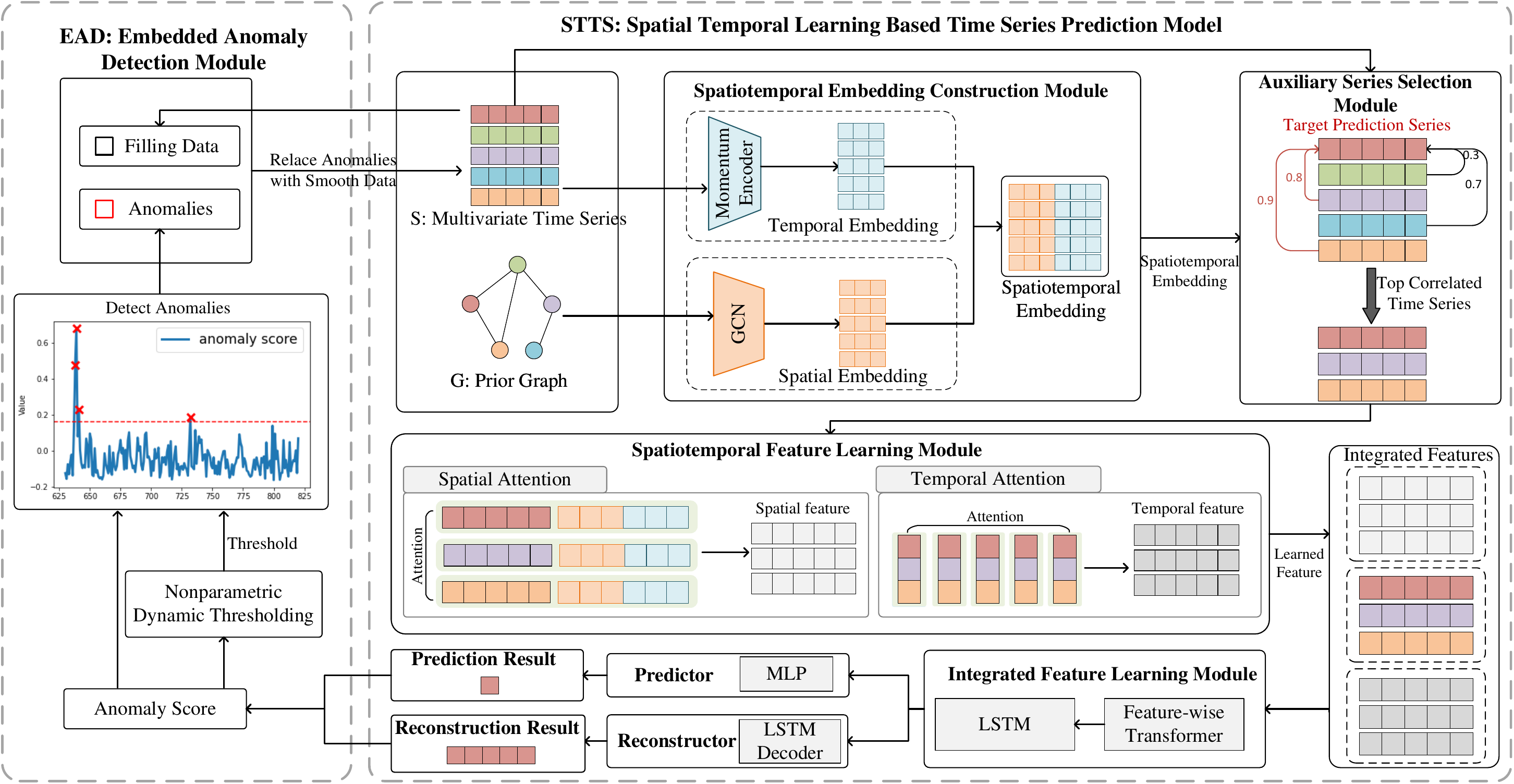}
    \caption{The architecture of STTS-EAD,
     with a spatio-temporal based prediction model and an anomaly detection module.
    }
    \label{fig:model}
\end{figure*}

\section{Method}

\subsection{Problem Statement and Overall Architecture}
\label{subsec:3.1}
As shown in Figure \ref{fig:model}, the raw MTS data is represented as $S \in \mathbb{R}^{N \times T}$, consists of $N$ individual time series $\{S_i \mid i = 1 \text{,...,} N\}$. The $N$ denotes the number of entities, also the feature dimension of the MTS data, and $T$ denotes the length of the MTS data. Typically, $N$ remains constant, but in certain scenarios, the number of entities dynamically changes over time. For instance, in the context of forecasting daily sales, each store represents an entity, and the daily sales of all stores form the MTS data. As the number of stores varies over time, the value of $N$ is not fixed.
To accommodate the dynamic evolution of the number of entities and harness inter-metric spatial information, we transform the raw MTS data into an input representation denoted as $X = \{X_{i,t} \mid 1 \leq i \leq N, 1 \leq t \leq T \}$.
$X_{i,t} \in \mathbb{R}^{N \times P}$ is processed as a sliding window input sample, where $P$ is the sliding window length, and $t$ and $i$ denote the timestamp and target prediction entity index, respectively. To construct $X_{i,t}$, the window data of the target prediction series $i$ is placed in the first dimension, with the remaining series concatenated as auxiliary series in subsequent dimensions to provide auxiliary spatial information.

If prior entity information is available, it can be used to create a spatial graph $\mathcal{G} \in \mathbb{R}^{N \times N}$ as an adjacency matrix, representing relationships among entities. For example, in sales forecasting, entities can be connected based on features like store category, city, or region. However, due to the flexibility and scalability of the proposed method, the model performs well even without prior spatial information. Thus, constructing $\mathcal{G}$ is beneficial but not essential.

The above formulation of the data aims to accomplish two goals as following:
\begin{itemize}
    \item {\textbf{Time Series Forecasting}}: works on both training and inference phase, given $X$ (and $\mathcal{G}$), aims to build a mapping $f(X, \mathcal{G}, \phi)$ to predict $Y$, the value of target time series at the next timestamp, $\phi$ is the model parameters. The formal representation is as follows:
        \begin{equation}
            \label{eq:prob}
            f(X,G,\phi) \to Y
        \end{equation}
    \item {\textbf{Anomaly Detection}}: only works during training to correct anomalies in the training data, given the input  $X$  (and $\mathcal{G}$), the real target value $Y$, the goal is to detect anomalies through a mapping $g(X, \mathcal{G}, Y, \phi)$.
    In the detection phase, the model parameters $\phi$ are constant.
        \begin{equation}
            \label{eq:prob_2}
            g(X, G, Y, \phi) \to C
        \end{equation}
\end{itemize}
During each iteration, only the training data is optimized for anomaly detection, with fixed model parameters. Conversely, when optimizing the prediction model, only the model parameters are updated, while the training data remains unchanged.
The STTS-EAD model consists of two main components: the Spatio-Temporal Learning-based Time Series prediction model (STTS) and the Embedded Anomaly Detection module (EAD), as shown in Figure \ref{fig:model}. The EAD module improves the STTS model by detecting and handling anomalies using residual information from training samples, replacing rough data preprocessing, enhancing training set quality, and refining model training. During training, the two components alternate to optimize each other, but during inference, only the STTS model is used for prediction.

As shown in Figure \ref{fig:model}, the STTS model includes several key components. The Spatiotemporal Embedding Construction Module captures temporal and spatial features of each time series. The Auxiliary Series Selection Module evaluates correlations between time series using spatiotemporal embeddings and selects relevant series as auxiliary inputs. The Spatiotemporal Feature Learning Module, including Temporal and Spatial Attention Modules, extracts high-dimensional temporal and spatial features. The Integrated Feature Learning Module refines the spatiotemporal information for the Predictor and Reconstructor. The Prediction Module uses the refined features for forecasting, while the Reconstructor performs reconstruction tasks. Errors from prediction and reconstruction are sent to the EAD module for anomaly detection.
The EAD module, activated periodically during training, detects and corrects anomalies in the training data to optimize model performance and improve predictive accuracy.

\subsection{Model Architecture of STTS}
The right part of Figure \ref{fig:model} shows the STTS model for time series prediction, consisting of key components: Spatiotemporal Embedding Construction, Auxiliary Series Selection, Spatiotemporal Feature Learning, Integrated Feature Learning, and a Predictor and Reconstructor. 
The Spatiotemporal Embedding Construction module combines temporal and spatial embeddings for each series. The Auxiliary Series Selection module uses these embeddings to select relevant time series, reducing computation and adapting to dynamic MTS data. 
The Spatiotemporal Feature Learning module captures temporal and spatial dependencies, while the Integrated Feature Learning module refines these features. 
The high-level features are then used by the Predictor for predictions and the Reconstructor for reconstruction, with errors provided to the EAD module for anomaly detection. 
Overall, STTS captures spatio-temporal patterns and produces accurate predictions.

\subsubsection{Spatiotemporal Embedding Construction Module}
The spatiotemporal embedding construction module constructs spatiotemporal embedding for each time series, which aims to learn the features of entities and ensures similar or related time series have similar embedding, while unrelated time series have distinct embedding.
The spatiotemporal embedding is concatenated by the temporal embedding $E_\text{time}\in \mathbb{R}^{d_1}$ learned from historical time series data and the spatial embedding $E_\text{spat} \in \mathbb{R}^{d_2}$ learned from prior graph information. $d_1$ and $d_2$ are dimensions of the corresponding embedding.
For the temporal embedding $E_\text{time}$ of each entity, a momentum encoder is employed to update it as follows:
\begin{equation}
    E_\text{time} = \gamma E_\text{time} + (1 - \gamma) f(X)
    \label{eq:node embedding}
\end{equation}
The former term uses the temporal embedding before iteration to represent the overall temporal characteristics and the latter uses an MLP layer $f$ to map local time window data $X$ to the local temporal characteristics and then adds the two parts with the fixed weight $\gamma$ to update the temporal embedding.
The weighted summation of local and global information allows the temporal embedding to not only incorporate short-term information from the current window but also retain long-term global sequential information accumulated from historical windows. And the incremental update approach enhances the stability and robustness of the temporal embedding construction. The $E_\text{time}$ is randomly initialized at first.

As for the spatial embedding construction, if prior graph knowledge $\mathcal{G}$ is available, $E_\text{spat}$ is encoded through a GCN \cite{kipf2016semi} network. The network aggregates the information of the connected neighbor entities, therefore spatial embedding can learn the dependency of similar and related entities. 
After the construction of temporal embedding and spatial embedding, the spatiotemporal embedding denoted as $E = [E_\text{time}, E_\text{spat}]\in \mathbb{R}^d, d=d_1 + d_2$ which formed by concatenating the temporal and spatial embedding. In addition, when prior graph knowledge $\mathcal{G}$ is not provided, the spatiotemporal embedding will solely consist of temporal embedding: $E = E_\text{time}$ and $d = d_1$. Excluding the calculation method of spatiotemporal embedding, the subsequent module operations will be identical regardless of whether $\mathcal{G}$ is provided or not.

\subsubsection{Auxiliary Series Selection Module}
The auxiliary series selection module selects relevant time series as external information for target series prediction and anomaly detection for three reasons. 
First, when the number of time series changes dynamically, this module selects a fixed number of series as input, allowing the model to adapt to dynamic MTS data. 
Second, when MTS data dimensionality is too high, predicting all series simultaneously becomes difficult. A common solution, as in \cite{hundman2018detecting}, is to model each series independently, but this approach doesn’t leverage spatio-temporal information from external series. Instead, we calculate the correlation between series using spatiotemporal embeddings and select the most relevant series to reduce computational costs and better utilize external spatio-temporal information.
Finally, for the anomaly detection module in \ref{subsec:EAD}, the single target series prediction approach eliminates the need for anomaly attribution across multiple time series.

This module calculates the correlation between other series $j$ and target series $i$ based on spatiotemporal embedding as follows:
\begin{equation}
    r_{ij} = \frac{E_i ^\top E_j}{\lVert E_i\rVert \cdot \lVert E_j\rVert}, j \in \{1,2,...,N\}
    \label{eq:node selection}
\end{equation}
The $r_{ij}$ is the correlation between two series. $E_i$ and $E_j$ are spatiotemporal embeddings of series $i$ and $j$ respectively, and $\lVert\cdot\rVert$ denotes vector norm. 
The module outputs the time series data of $M$ series (including the target prediction series itself) with the highest similarity to the target series $i$, represented as $X' \in \mathbb{R}^{M \times P}$ based on the similarity $r_{ij}$. $M$ is a manually selected hyperparameter.

\subsubsection{Spatiotemporal Feature Learning Module}
The spatiotemporal feature learning module comprises a temporal attention block and a spatial attention block, designed to effectively capture and model both intra-metric and inter-metric dependencies with attention mechanisms. 
After filtered by the auxiliary series selection module, $X'$ is input into the spatiotemporal feature learning module.
For the input $X'$, it can be considered as a matrix of shape $M \times P$, where each row corresponds to the time series window data of a series ($M$ series in total) and each column represents cross-sectional features of a timestamp ($P$ timestamps in total). We denote the $i$-th row as $X'_{i,:} \in \mathbb{R}^P$, representing the $i$-th time series, and the $i$-th column represents the $i$-th timestamp feature denoted as $X'_{:,i} \in \mathbb{R}^M$.
For Spatiotemporal Feature Learning, GATv2 \cite{brody2021attentive} is employed to learn the temporal representations $H^{\text{time}} \in \mathbb{R}^{M \times P}$ and spatial representation $H^{\text{spat}}  \in \mathbb{R}^{M \times P}$.
These representations are obtained via weighted summation with attention scores $\alpha_{ij}$. 

In the temporal attention block, each computing unit corresponds to a column vector $X'_{:, i}$ as the equations \eqref{eq:h_time}.
\begin{equation}
\begin{aligned}
    & H_i^{\text{time}} = \sigma(\sum\limits_{j=1}^{P} \alpha_{ij}X_{:,j}') \\
    & p_{ij} = \mathbf{a}^\top \cdot \text{LeakyReLU}(\mathbf{W}\cdot(X_{:,i}' \oplus X_{:,j}')) \\
    & \alpha_{ij} = \frac{exp(p_{ij})}{\sum_{k=1}^P(exp(p_{ik}))}
\end{aligned}
\label{eq:h_time}
\end{equation}
In the representations calculation, $\mathbf{a}\in\mathbb{R}^{d'}$ and $\mathbf{W} \in \mathbb{R}^{d'\times 2d_u}$ are learned parameters, $\text{LeakyReLU}$ is the activate function, and $\oplus$ is the concatenate operation. And $d'$ is the manually set dimension of the intermediate hidden state, $d_u$ corresponds to the dimension of the attention computing unit. In temporal attention calculation, $d_u = M$. While in spatial attention calculation, $d_u = P + d$. Because each computing unit is a concatenation of a row vector $X'_{i,:}$ and the corresponding spatiotemporal embedding $E_i$ as described by the equations \eqref{eq:h_spat}. The spatiotemporal embedding contains rich spatiotemporal information about the entity, enriching the learning of spatial representations.
\begin{equation}
\begin{aligned}
    & H_i^{\text{spat}} = \sigma(\sum\limits_{j=1}^{M} \alpha_{ij}(X'_{j,:} \oplus E_j)) \\
    & p_{ij} = \mathbf{a}^\top \cdot \text{LeakyReLU}(\mathbf{W}\cdot((X_{i,:}' \oplus E_i) \oplus (X_{j,:}' \oplus E_j))) \\
    & \alpha_{ij} = \frac{exp(p_{ij})}{\sum_{k=1}^M(exp(p_{ik}))}
\end{aligned}
\label{eq:h_spat}
\end{equation}

To prevent information loss, the selected series $X'$ is concatenated with the two representations $H^{\text{time}}$ and $H^{\text{spat}}$, resulting in $\tilde{X} \in \mathbb{R}^{3M \times P}$ for input into subsequent modules.

\subsubsection{Integrated Feature Learning Module}
The integrated feature learning module consists of a feature-wise transformer block and an LSTM block. After integrating the selected series, temporal, and spatial representations into $\tilde{X}$, the feature-wise transformer block uses self-attention to learn complex relationships between the representations and auxiliary series. This step enhances the blending of information and better captures spatial dependencies. 
The output from the transformer block is then passed to the LSTM layer, which learns temporal patterns and characteristics. The resulting hidden state captures important high-level spatio-temporal information.

\subsubsection{Predictor \& Reconstructor}
STTS-EAD consists of two output layers: a predictor and a reconstructor. By jointly optimizing these two layers, STTS-EAD is able to accurately predict target values while providing residual information to the EAD module for anomaly detection in training samples. The predictor, which is a fully connected layer is designed to predict target values and generate prediction errors for the EAD module. The reconstructor is an LSTM decoder that aims to reconstruct the target sequence within the sliding window, providing reconstruction errors for EAD. The overall loss function is the weighted sum of two optimization targets:
\begin{equation}
    \label{eq:obj_2}
    \begin{aligned}
        & L = \beta L_{\text{pred}} + (1 - \beta) L_{\text{rec}} \\
        & L_{\text{pred}} = \sqrt{(Y - \hat{Y})^2} \\
        & L_{\text{rec}} = \sqrt{\frac{1}{P}\sum_{t=1}^P(X_{\text{target},t} - \hat{X}_{\text{target},t})^2}
    \end{aligned}
\end{equation}
Here, $\beta$ serves as the weight coefficient for two parts of losses. $L_{\text{pred}}$ represents the loss from predicting the target sequence, quantifying the discrepancy between the predicted $\hat{Y}$ and the actual prediction label $Y$. $L_{\text{rec}}$ denotes the reconstruction loss for the target series window, reflecting the deviation between the reconstructed sliding window of the target prediction series $\hat{X}_{\text{target},t:t+P}$ and the raw window data $X_{\text{target},t:t+P}$. Both losses are computed using the Root Mean Square Error metric.

\subsection{Improve Training with EAD module}
\label{subsec:EAD}
\IncMargin{1em}
\begin{algorithm} \SetKwData{Left}{left}\SetKwData{This}{this}\SetKwData{Up}{up} \SetKwFunction{Union}{Union}\SetKwFunction{FindCompress}{FindCompress} \SetKwInOut{Input}{Input}\SetKwInOut{Output}{Output}
	
	\Input{Input data $S \in \mathbb{R}^{N \times T}$, training iterations $n_{epoch}$, EAD execution period $\eta$.} 
	\Output{Trained STTS-EAD model}
	 
	 Preprocess data $S^{(0)}$ and initialize data loader $D^{(0)}$\;
	 \For{$e\gets 0$ \KwTo $(n_{epoch} -1 )$}{ 
	 	\text{Train STTS model with data loader} $D^{(e)}$\;
            \If{$e \% \eta = 0$}{
                Get prediction error \{$\epsilon_p^1,...,\epsilon_p^{N\times T}\}$ and reconstruction error $\{\epsilon_r^1,...,\epsilon_r^{N\times T}\}$ from STTS;
                
                Get anomaly scores $\{s^i \gets \delta \epsilon_p^i + (1-\delta)\epsilon_r^i\}_{i=1}^{N\times T}$\;

                $\tau \gets$ Calculate threshold with scores $\{s^i\}_{i=1}^{N\times T}$;

                $A \gets$ Get the position of anomalies where $s^i > \tau$\;

                $S^{(e+1)}\gets$ Replace anomalies in set $A$ with smoothing fill data in $S^{(e)}$\;

                $D^{(e+1)} \gets$ Reconstruct data loader with $S^{(e+1)}$\;
            }
            \Else{$S^{(e+1)}, D^{(e+1)} \gets S^{(e)}, D^{(e)}$\;} 
 	 }
   \Return \text{Trained STTS-EAD Model}
      \caption{STTS-EAD training algorithm}
      \label{algo_disjdecomp} 
 \end{algorithm}
 \DecMargin{1em}
Our core idea in STTS-EAD is the Embedded Anomaly Detection (EAD) module in the training phase aims to detect and rectify the anomalies in the MTS data dynamically to clean the training dataset, thereby optimizing the training of the model, and improve the prediction performance of the STTS model. The training procedure of STTS with the EAD module is shown in Algorithm \ref{algo_disjdecomp}.
The dataset $S_0$ is loaded into $D_0$ for model training. Every $\eta$ epochs, the EAD procedure runs (Algorithm \ref{algo_disjdecomp}, lines 4-10). It calculates prediction error $e_p$ and reconstruction error $e_r$, generating anomaly scores $s$ weighted by $\delta$. A dynamic thresholding method \cite{hundman2018detecting} determines the threshold $\tau$, and points with scores above $\tau$ are marked as anomalies (set $A$). These anomalies are replaced with filling data (Section \ref{subsubsec:filling_data}) to improve data quality. Training continues with the modified dataset $S^{(e)}$ until $n_{\text{epoch}}$ iterations are completed. Hyperparameters include $n_{\text{epoch}}$, $\eta$, and $\delta$.


\section{Experiments and Analysis}
\subsection{Experimental Setup}

\noindent  \textbf{Datasets}: Three datasets are used to validate the effectiveness of the proposed model: Coffee-Bean (espresso roast coffee bean sales), Coffee-Cream (cream sales in coffee drinks), and Stock-SP500 (closing prices of the S\&P 500). Detailed statistics for each dataset are provided in Table \ref{tab:dataset}, including the number of series, timestamp length, sample counts for the training, validation, and testing sets, and whether the dataset includes prior spatial information $\mathcal{G}$.

\begin{table}
  \centering
  \caption{Dataset statistics.)}
  \label{tab:dataset}
  \tabcolsep=0.1cm
  \begin{tabular}{lcccccc}
    \toprule
    Datasets & \#Series & \#Timestamps & Train & Valid & Test & $\mathcal{G}$ \\
    \midrule
    Coffee-Bean & 500 & 1096 & 400500 & 36000 & 82000 & \checkmark \\
    Coffee-Cream & 500 & 1096 & 400500 & 36000 & 82000 & \checkmark \\
    Stock-SP500 & 483 & 1258 & 444843 & 41055 & 93219 & $\times$ \\
  \bottomrule
\end{tabular}
\end{table}


    \noindent \textbf{Baseline Methods}: We compared STTS-EAD with nine time series forecasting baseline models of six categories, i.e., traditional univariate model: ARIMA \cite{box2015time}, RNN-based model: LSTM \cite{hochreiter1997long}, three transformer-based models: Informer \cite{zhou2021informer}, Autoformer \cite{wu2021autoformer} and Preformer \cite{du2023preformer}, CNN-based model: TCN \cite{bai2018empirical}, MLP-based model: DLinear and NLinear \cite{zeng2023transformers} and GNN-based model: MTGNN \cite{wu2020connecting}.

\noindent  \textbf{Evaluation}: Prediction accuracy is evaluated using RMSE and MAE metrics.

\subsection{Prediction Performance Evaluation}
\begin{table*}
    \caption{The time series forecasting results of various methods on three datasets, including Prediction Performance results, ablation study for STTS, and EAD module evaluation results.}
    \label{tab:performance}
    \setlength{\tabcolsep}{14pt}
    \renewcommand{\arraystretch}{1.0}
    \begin{tabular}{lcccccc}
        \toprule
        \multirow{2}*{Model} & \multicolumn{2}{c}{Coffee-Bean} &  \multicolumn{2}{c}{Coffee-Cream}  & \multicolumn{2}{c}{Stock-S\&P500}\\ 
        \cline{2-7}
        & RMSE & MAE & RMSE & MAE & RMSE & MAE \\
        \cline{1-7}
        \multicolumn{7}{c}{Part 1: Prediction Performance Evaluation} \\
        \cline{1-7}
        ARIMA \cite{box2015time} & 1.68247 & 1.17438 & 1.89212 & 1.35280 & 0.03051 & 0.02177 \\
        LSTM \cite{hochreiter1997long}  & 1.00314 & 0.71841 & 1.22049 & 0.87284 & 0.06078 & 0.05836 \\
        TCN \cite{bai2018empirical}  & 1.38165 & 1.06422 & 1.33837 & 0.93443 & 0.04556 & 0.04266 \\
        MTGNN \cite{wu2020connecting} & \underline{0.96443} & \underline{0.65518} & 1.13830 & 0.77632 & 0.01384 & 0.01029 \\
        Informer \cite{zhou2021informer} & 1.09768 & 0.76968 & 1.38418 & 1.06042 & 0.05220 & 0.04740 \\
        Autoformer \cite{wu2021autoformer} & 0.97758 & 0.69155 & 1.38381 & 0.95716 & 0.01517 & 0.01111 \\
        Preformer \cite{du2023preformer} & 1.20185 & 0.89936 & \underline{1.13533} & \underline{0.77578} & 0.01584 & 0.01186 \\
        DLinear \cite{zeng2023transformers} & 1.05390 & 0.73760 & 1.19470 & 0.83830 & 0.01416 & 0.00996 \\
        NLinear \cite{zeng2023transformers} & 1.06299 & 0.74458 & 1.21750 & 0.83861 & \underline{0.01331} & \underline{0.01013} \\
        \cdashline{1-7}[3pt/2pt]
         STTS & 0.92989 & 0.63513 & 1.11719 & 0.75777 & 0.01231 & 0.00842 \\
        Gain (\%) & 3.6\% & 3.1\% & 1.6\% & 1.3\% & 7.5\% & 16.9\% \\
        \cline{1-7}
        {\bf STTS-EAD}  & {\bf 0.91004} & {\bf 0.62196} & {\bf 1.09157} & {\bf 0.74425} & {\bf 0.01223} & {\bf 0.00834} \\
        Gain (\%) & 5.6\% & 5.1\% & 3.8\% & 4.0\% & 8.1\% & 17.7\% \\
        \cline{1-7}
        \multicolumn{7}{c}{Part 2: Ablation Study for STTS} \\
        \cline{1-7}
        STTS & 0.92989 & 0.63513 & 1.11719 & 0.75777 & 0.01231 & 0.00842\\
        \cdashline{1-7}[3pt/2pt]
        w/o Auxiliary Entity Selection & 0.96615 & 0.65969 & 1.14410 & 0.78349 & 0.01293 & 0.00928 \\
        w/o Spatial Attention & 0.93107 & 0.64644 & 1.13332 & 0.77391 & 0.01244 & 0.00864 \\
        w/o Temporal Attention & 0.93299 & 0.63845 & 1.12801 & 0.77083 & 0.01241 & 0.00854 \\
        w/o Feature-wise Transformer & 0.95821 & 0.65469 & 1.13775 & 0.77697 & 0.01239 & 0.00858 \\
        w/o LSTM & 0.94122 & 0.65593 & 1.16964 & 0.81458 & 0.01238 & 0.00845 \\
        \cline{1-7}
        \multicolumn{7}{c}{Part 3: EAD Module Evaluation: Replaced with Other Anomaly Detection Preprocessing} \\
        \cline{1-7}
        STTS & 0.92989 & 0.63513 & 1.11719 & 0.75777 &  0.01231 & 0.00842 \\
        \cdashline{1-7}[3pt/2pt]
        \textbf{STTS-EAD} & \textbf{0.91004} $\uparrow$ & \textbf{0.62196} $\uparrow$ & \textbf{1.09557} $\uparrow$ & \textbf{0.74919} $\uparrow$ & \textbf{0.01223} $\uparrow$ & \textbf{0.00834} $\uparrow$ \\
        STTS-3$\sigma$ & 0.94088 $\downarrow$ & 0.64627 $\downarrow$ & 1.11522 $\uparrow$ & 0.76390 $\downarrow$ & 0.01233 $\downarrow$ & 0.00841 $\uparrow$ \\
        STTS-EWMA \cite{carter2012probabilistic} & 0.93992 $\downarrow$ & 0.64514 $\downarrow$ & 1.12617 $\downarrow$ & 0.77791 $\downarrow$ & 0.01238 $\downarrow$ & 0.00843 $\downarrow$ \\
        STTS-USAD \cite{audibert2020usad} & 0.95143 $\downarrow$ & 0.64604 $\downarrow$ & 1.11669 $\uparrow$ & 0.78785 $\downarrow$ & 0.01235 $\downarrow$ & 0.00842 $-$ \\
        STTS-LSTM-NDT \cite{hundman2018detecting} & 0.92576 $\uparrow$ & 0.63307 $\uparrow$ & 1.11395 $\uparrow$ & 0.76559 $\downarrow$ & 0.01231 $-$ & 0.00839 $\uparrow$ \\
        \bottomrule
    \end{tabular}
\end{table*}
The evaluation results in Part 1 of Table \ref{tab:performance} show that bold entries correspond to the best results, achieved by the STTS-EAD method. These results demonstrate that STTS-EAD significantly outperforms all strong baseline methods. Compared to the baselines, \textbf{STTS-EAD} shows improvements of 5.6\%, 3.8\%, and 8.1\% in RMSE on three datasets, respectively. Additionally, \textbf{STTS} without the EAD anomaly detection module still improves by 3.6\%, 1.6\%, and 7.5\%. These findings highlight that STTS leverages spatiotemporal information for accurate predictions, and the EAD module further enhances performance.

Specifically, (1) ARIMA performed poorly because it doesn't capture external information. (2) The TCN model struggled with modeling correlations between variables and handling high-dimensional features. (3) The LSTM model was strong on two sales datasets but weak on the Stock-SP500 dataset, likely due to differences in stationarity and predictability. (4) The Autoformer outperformed the Informer model, thanks to its Auto-Correlation mechanism. (5) The MLP-based methods, NLinear and DLinear, were competitive but limited in learning spatial relationships between series. (6) MTGNN, which constructs a graph for time series and uses graph convolution for feature extraction, performed second best after STTS-EAD, excelling with high-dimensional data and external graph data.



\subsection{Ablation Study for STTS}
\label{subsec:ablation}
We conduct ablation experiments on the STTS method to assess the role of each component. The second part of Table \ref{tab:performance} shows that removing any component increases prediction errors, proving their importance. Without the auxiliary series selection module, errors increase by 3.6\% and 5.4\%, highlighting its role in selecting relevant series. Removing the spatial attention module increases errors by 0.9\% and 2.1\%, and removing the temporal attention module increases them by 0.9\% and 1.0\%, showing the value of both attention mechanisms. Removing the Feature-wise Transformer and LSTM layers causes RMSE to increase by 1.8\% and 2.1\%, and MAE by 2.4\% and 3.5\%, emphasizing their necessity.

\subsection{EAD Evaluation and Analysis}
\subsubsection{Comparison of EAD with other anomaly detection approaches}
\label{subsubsec:comp_EAD}

To further evaluate EAD, we replaced it with traditional "two-stage" anomaly preprocessing methods while keeping the STTS model unchanged. We used different anomaly detection methods for data preprocessing, and the processed data was then used to train and evaluate the STTS model.
We compared two statistical methods—3$\sigma$ and Exponentially Weighted Moving Average (EWMA)—and two deep learning-based methods—USAD and LSTM-NDT.
The results in Part 3 of Table \ref{tab:performance} show that the model with EAD outperforms all baselines, proving its ability to handle anomalies and improve forecasting. Some preprocessing methods improve performance, while others degrade it, indicating they fail to distinguish between anomalous and normal data. In contrast, the STTS-EAD model optimizes both anomaly detection and training, leading to better results.

\subsubsection{Impact of Anomaly Score Weight $\delta$}
\begin{table}
    \caption{The Impact of Parameter $\delta$ on Prediction Accuracy in the Coffee-Bean Dataset}
    \label{tab:4-delta}
    \centering
    \begin{tabular}{lcccccc}
        \toprule
        \multirow{2}*{$\delta$} & \multicolumn{2}{c}{Coffee-Bean} &  \multicolumn{2}{c}{Coffee-Cream}  & \multicolumn{2}{c}{Stock-SP500}\\ 
        \cline{2-7}
        & RMSE & MAE & RMSE & MAE & RMSE & MAE \\
        \cline{1-7}
        0 & 0.93844 & 0.64468 & 1.10239 & 0.75011 & 0.01236 & 0.00851 \\
        0.2 & 0.92521 & 0.62524 & 1.10140 & 0.74950 & 0.01230 & 0.00842 \\
        0.5 & \textbf{0.91004} & \textbf{0.62196} & \textbf{1.09557} & \textbf{0.74919} & 0.01226 & 0.00836 \\
        0.8 & 0.92586 & 0.63631 & 1.10870 & 0.75310 & \textbf{0.01223} & \textbf{0.00834} \\
        1.0 & 0.92632 & 0.62903 & 1.11114 & 0.75808 & 0.01229 & 0.00842 \\
        \bottomrule
    \end{tabular}
\end{table}

In anomaly detection, the anomaly score measures the abnormality of data points, helping algorithms identify potential anomalies. Higher scores indicate greater abnormality. In Algorithm \ref{algo_disjdecomp}, the score is calculated as the weighted sum of prediction error $e_p$ and reconstruction error $e_r$. To explore their impact on the score, we adjust the $\delta$ parameter to change the weights of these errors.

The results in Table \ref{tab:4-delta} show how different $\delta$ values affect the score. As $\delta$ increases, the weight of prediction error rises, while reconstruction error's weight decreases. In both sales datasets, the prediction error is minimized at $\delta = 0.5$, while in the Stock-SP500 dataset, it is minimized at $\delta = 0.8$. The results also reveal that using either error alone is less effective than combining them. This combined approach more accurately identifies anomalies, as the predictor is sensitive to time series randomness, and the reconstructor is robust to noise. This complementarity improves performance in complex data environments.

\subsubsection{Effectiveness of Filling Data}
\label{subsubsec:filling_data}
\begin{figure}
    \centering
    \includegraphics[width=\linewidth]{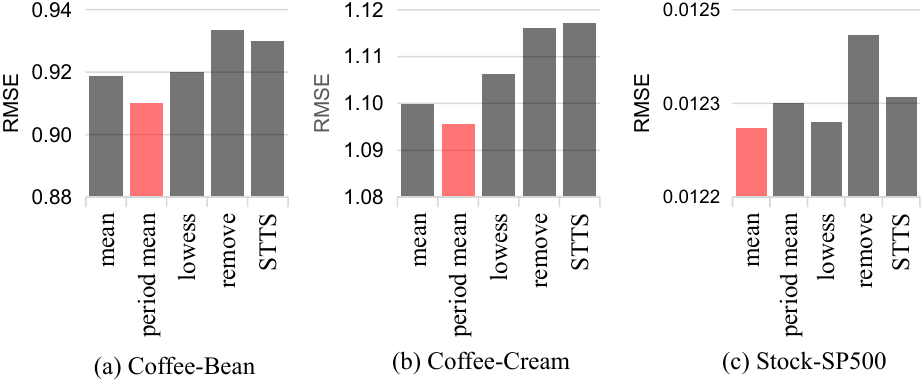}
    \caption{Performance with different filling Data. (RMSE)}
    \label{fig:aux2}
\end{figure}
    
During the EAD process, accurately filling detected anomalies with appropriate data is essential. The filling data should match the normal data distribution; otherwise, it may not improve data quality and could have negative effects. Figure \ref{fig:aux2} compares various methods for filling anomaly locations. Most methods improve multivariate time series prediction compared to the initial STTS. However, removing anomalies is less effective than the other three methods, which use smooth data filling. The "periodic mean" filling method works better for coffee sales datasets, while "mean" and "lowess" methods perform better for stock datasets. These differences arise from the distinct characteristics of the data, with coffee sales showing strong cyclical correlation and stock data focusing on nearby point distributions.

\subsubsection{Case Study on EAD module}
\begin{figure}
    \centering
    \includegraphics[width=\linewidth]{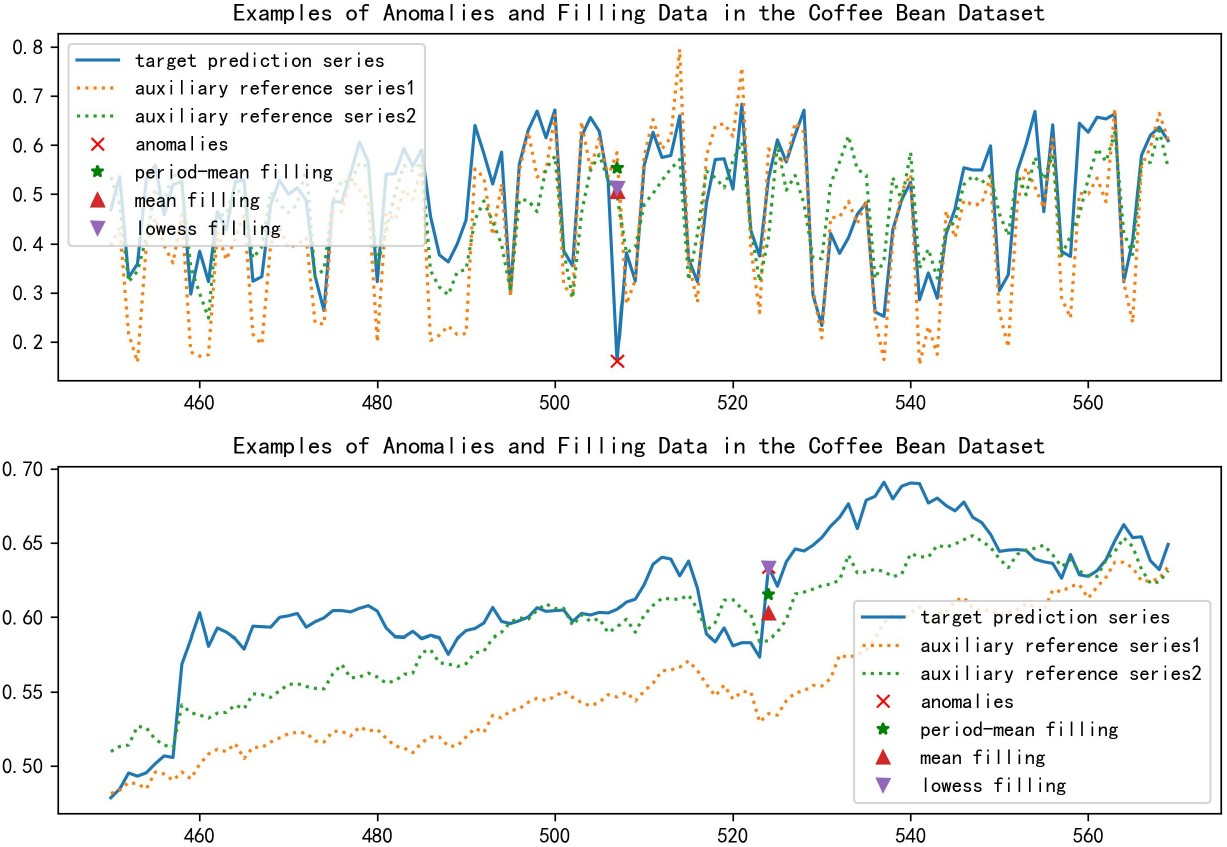}
    \caption{Case Study on anomaly detection and filling in the EAD module.}
    \label{fig:Ano_Visual}
\end{figure}

Figure \ref{fig:Ano_Visual} demonstrates the effectiveness of the EAD module in anomaly detection and data imputation on the Coffee-Bean and Stock-SP500 datasets after 10 training rounds. The figure includes the target prediction series, two reference series from the auxiliary sequence selection module, anomalies marked by red crosses, and three data filling methods.
In the Coffee-Bean dataset, both target and reference series show clear periodic patterns. A downward spike anomaly is identified due to its inconsistency with historical data, even though it doesn't deviate much from the overall distribution. Mean and lowess filling methods result in smaller replacement values than the period-mean filling, as they rely on neighboring data, while period-mean filling uses historical data from the same period for more accurate results.
In the Stock-SP500 dataset, where sequences are less consistent, mean filling better approximates the normal data distribution. A sharp upward trend anomaly, likely caused by a sudden event, is identified and adjusted by mean filling to a more reasonable level.

\section{Conclusion}
We propose a novel approach, STTS-EAD, designed to improve the accuracy of time-series prediction based on spatio-temporal learning using an embedded anomaly detection module that dynamically detects anomalies during the training phase to optimize the training process.
Our experiments demonstrated the superior performance of our STTS-EAD model and the significant improvement in prediction accuracy achieved by using the EAD module.


%





\ifCLASSOPTIONcaptionsoff
  \newpage
\fi



\bibliographystyle{IEEEtran}
\bibliography{sample-base.bib}
\end{document}